\documentclass[runningheads]{llncs}
\usepackage{graphicx}
\usepackage{amsmath,amssymb} 
\usepackage{color}
\usepackage[width=122mm,left=12mm,paperwidth=146mm,height=193mm,top=12mm,paperheight=217mm]{geometry}
\usepackage[ruled, linesnumbered]{algorithm2e}
\usepackage{multirow}
\usepackage{hyperref}
\usepackage{xcolor}
\hypersetup{
	colorlinks   = true, 
}
\DeclareMathOperator*{\argmin}{arg\,min}
\newcommand{\etal}{\textit{et al} }

\begin{document}
\pagestyle{headings}
\mainmatter

\title{Distillation Techniques for Pseudo-rehearsal Based Incremental Learning} 

\titlerunning{Distillation Techniques for Pseudo-rehearsal Based Incremental Learning}

\authorrunning{Haseeb Shah, Khurram Javed and Faisal Shafait}

\author{Haseeb Shah}

\institute{Department,\\
	University\\
	\email{ \{author1,author2\}@univ.edu}
}

\author{Haseeb Shah, Khurram Javed and Faisal Shafait}

\institute{School of Electrical Engineering and Computer Science,\\
	National University of Science and Technology,\\
	\email{ \{hshah.bese15seecs, 14besekjaved, faisal.shafait\}@seecs.edu.pk}
}

\maketitle

\begin{abstract}
	The ability to learn from incrementally arriving data is essential for any life-long learning system. However, standard deep neural networks forget the knowledge about the old tasks, a phenomenon called catastrophic forgetting, when trained on incrementally arriving data. We discuss the biases in current Generative Adversarial Networks (GAN) based approaches that learn the classifier by knowledge distillation from previously trained classifiers. These biases cause the trained classifier to perform poorly. We propose an approach to remove these biases by distilling knowledge from the classifier of AC-GAN. Experiments on MNIST and CIFAR10 show that this method is comparable to current state of the art rehearsal based approaches. The code for this paper is available at this \href{https://github.com/haseebs/Pseudo-rehearsal-Incremental-Learning}{link}.
\end{abstract}

\section{Introduction} \label{intro}

Standard deep neural networks can solve a variety of simple and complex problems.
Most commonly nowadays, these networks are trained on data that is collected beforehand. This is not feasible in some applications, such as systems that require life-long learning. In such cases, the network has to be trained on incrementally arriving data. However, current standard neural networks tend to forget the knowledge they obtained from previous training sessions when trained incrementally. This phenomenon is known as catastrophic forgetting.\medskip

\noindent One particular way of overcoming forgetting is to retrain the network on old training 
examples in addition to the new training examples \cite{icarl}. These examples can 
be obtained either by storing the exemplars from previous increments, or storing a GAN
\cite{Goodfellow} that generates these examples. We empirically study the the latter
approach, using Auxiliary Classifier GAN (AC-GAN) \cite{acgan} in particular.\medskip

\noindent We find that the samples produced by GANs on simple datasets
like MNIST are accurate enough for retaining knowledge. However, we observe that distilling
knowledge from the old classifier when incrementally training on new classes 
results in a biased classifier. We propose a way to remove this bias by instead using AC-GAN
to generate images and using the auxiliary classifier of the discriminator for distillation.\medskip

\noindent The rest of this paper is organized as follows. In Section \ref{prelim}, we introduce GANs, knowledge
distillation and NCM classifier. In Section \ref{modeldist}, we discuss model distillation. In Section \ref{modeldistbias}, 
we discuss the biases introduced in the classifier when using model distillation. In Section \ref{acdist},
we discuss the use of AC-Distillation to remove these biases. In Section \ref{related}, we discuss the
related work on this topic and the baselines. In Section \ref{experiments}, we perform empirical analysis of
the approach, analyze the bias and discuss limitations. Finally in Section \ref{conclusion}, we conclude the paper.

\begin{figure}[t]
	\centering
	\begin{tabular}{c@{\hspace{1.2cm}}c}
		\includegraphics[height=7cm]{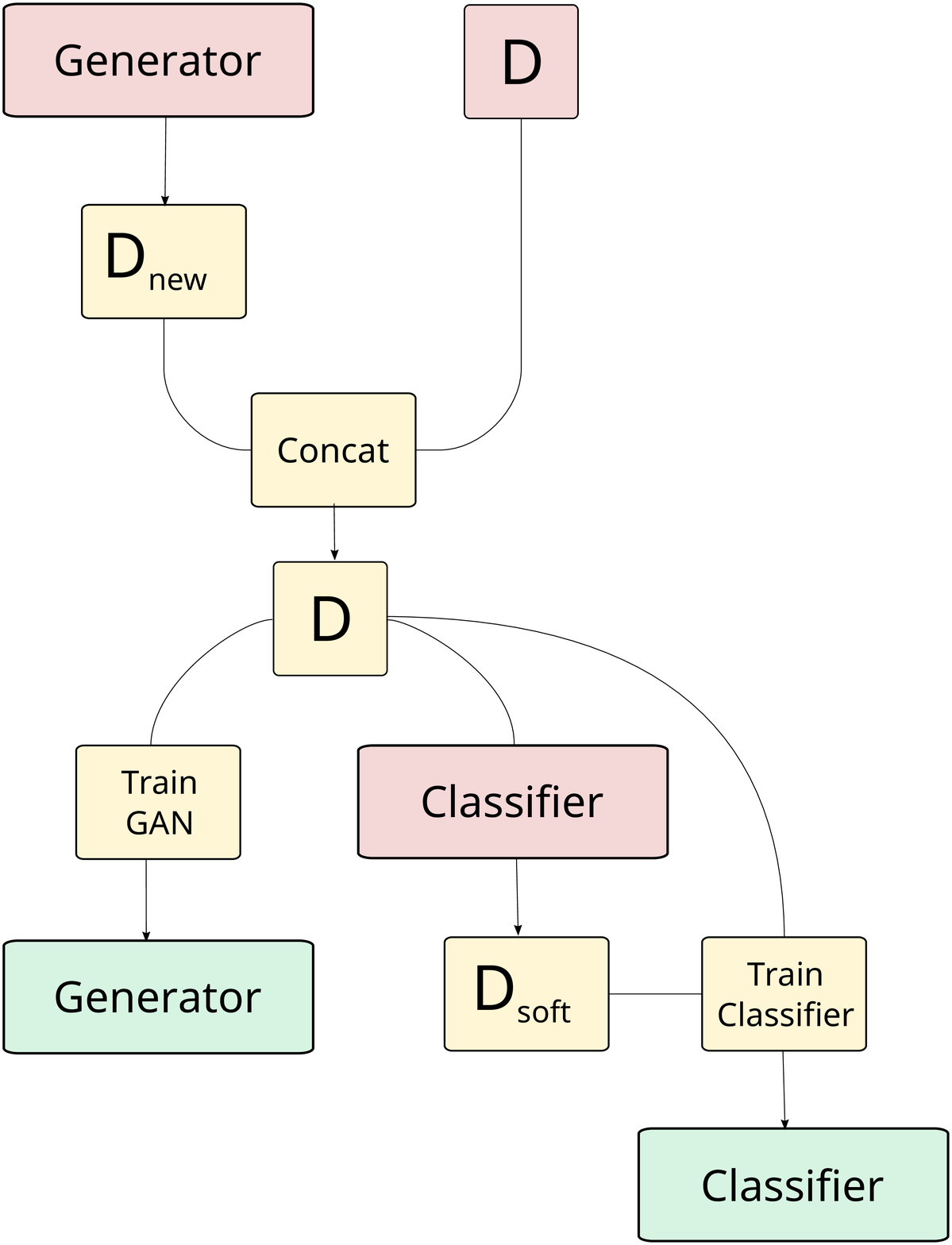}&
		\includegraphics[height=7cm]{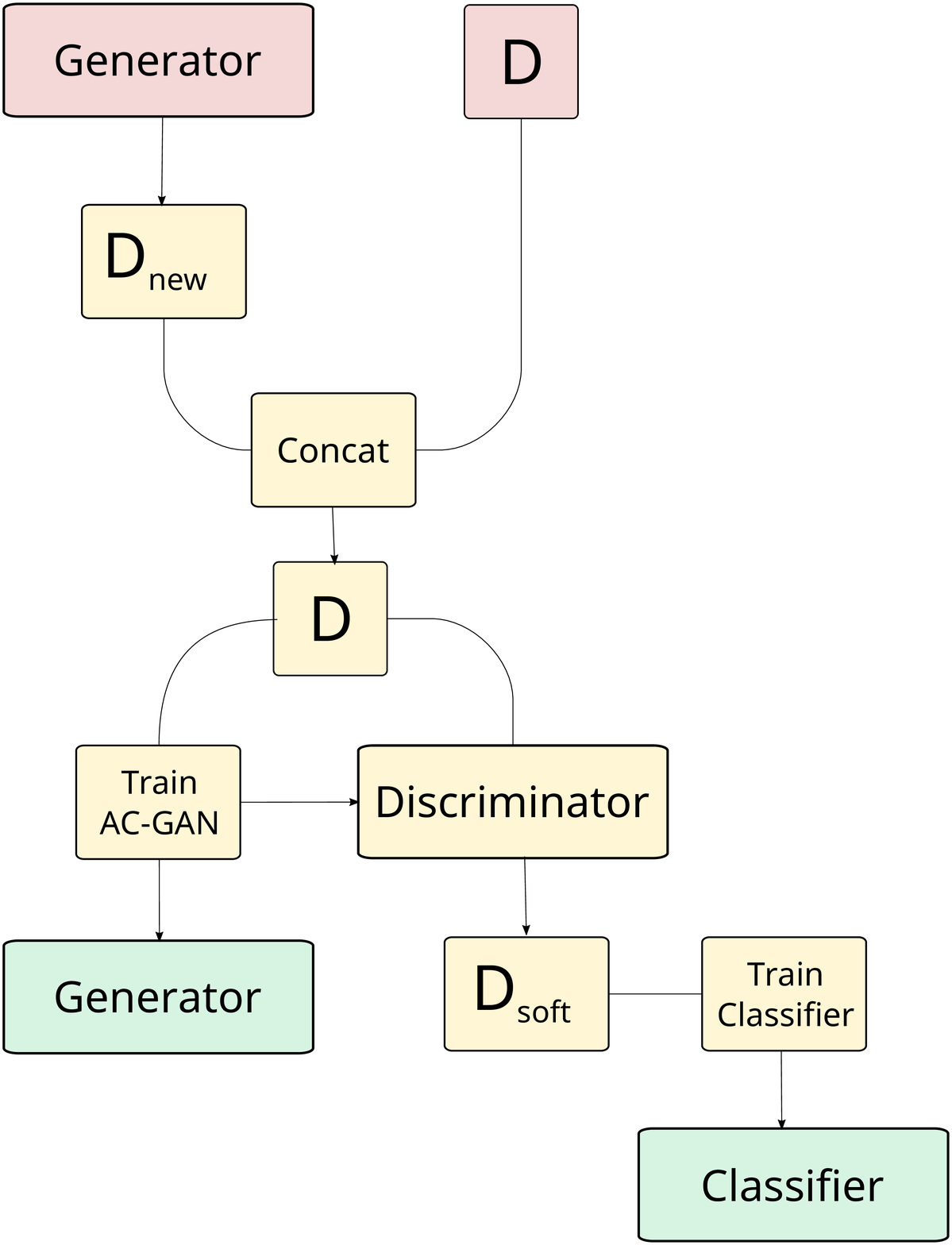}\\
		(a) Model Distillation &(b) AC-Distillation
	\end{tabular}
	\caption{(a) Architecture for the Model Distillation process as described in Section \ref{modeldist}. (b) Architecture for AC Distillation process described in Section \ref{acdist}. Inputs of each learning session are highlighted as red while the outputs are highlighted green. At each increment, these inputs are fed into the algorithm. Firstly, the new dataset $\mathcal D$ is augmented with the data generated belonging to old classes. Secondly, we train a new GAN using this dataset $\mathcal D$. In Model Distillation, the old classifier is used to generate soft targets which are then used in combination with real targets to train a new classifier. Whereas in AC Distillation, the discriminator of AC-GAN is used to generate the soft targets which are then used directly to train the classifier.}
	\label{fig:teaser}
\end{figure}


\section{Preliminaries} \label{prelim}
\subsection{Generative Adversarial Networks (GANs)}
Generative Adversarial Networks, introduced by Goodfellow 
\etal \cite{Goodfellow}, is a framework used to mimic a target distribution. GANs 
consist of a generator, which generates the samples that mimic the target
distribution and a discriminator, that gives the probability distribution over
sources. Both these networks are trained in turns, with discriminator acting as a
critic of increasing difficulty for the generator.

Conditional GANs (cGANs) are an extension of GANs \cite{cgan} which additionally utilise
conditional information in order to learn the target distribution.
These are used for class conditional image generation \cite{cgan}, text to
image generation \cite{ttoi} and image manipulation \cite{itoi}. One particular method of
utilising the conditional information, known as Auxiliary Classifier Generative
Adversarial Networks (AC-GAN) \cite{acgan}, incorporates a classifier in discriminator.
This classifier allows AC-GAN's discriminator to output the class probabilities in addition to 
probabilities of the source of input.\\

\subsection {Knowledge Distillation} 
Knowledge distillation is a technique that can be used to transfer knowledge
between neural networks. This transfer is achieved by training the target
network using output probabilities of source network as target probabilities.
These targets have higher entropy and thus contain similarity information
that can be utilised while training. On simple datasets like MNIST, the 
probabilities of non-target class are often very small. Caruana \etal \cite{caruana}
minimized the squared distance between logits to deal with this problem.
Hinton \etal \cite{hinton} propose a more general solution by raising the temperature
of softmax to produce softer targets and use them directly.\\

\subsection {Nearest Class Mean (NCM) Classifier} 
Nearest Class Mean classifier \cite{ncm} is a classification method where instead of 
using the trained classifier to perform the classification directly, the class of an 
image is picked as the one whose embeddings have minimum euclidean distance from the
mean embeddings of all images in that class. More formally,
\begin{equation}
c^* = \argmin_{c=1\dots\,t}||\phi(x) - \mu(c)||
\end{equation}
Where $c$ is the class, $\phi$ is the trained classifier and $\mu$ is the true class
mean calculated using all the examples present in $c$. Rebuffi \etal \cite{icarl} proposed 
{\it Nearest Mean of Exemplars} rule where instead of calculating the true class mean, the mean 
$\mu$ is calculated on a smaller set of exemplars acting as a representative of class $c$.


\section{Related Work} \label{related}
In this section, we discuss the work related to our proposed approach in the literature. \medskip

\noindent McCloskey \& Cohen \cite{mccloskey} explored the use of rehearsal in order to prevent forgetting. During rehearsal, a random set of training examples from old training sessions are picked and mixed with the ones in new training session. Robins \cite{robins} proposed pseudorehearsal, in which forgetting is prevented by generating random examples and then passing them through the network to obtain their labels. These examples along with their labels are then used during rehearsal.\medskip

\noindent Li and Hoiem \cite{lwf} introduce a method named Learning without Forgetting (LwF) wherein the distillation loss is used to prevent catastrophic forgetting. This method preserves the privacy as previous class data from previous training sessions is not stored. Rebuffi \etal \cite{icarl} introduced iCarl, which involves storing a representative set of exemplars for classes encountered in old training sessions using a process called herding. At each increment, they train the classifier using distillation and the classification losses combined. During classification, they use Nearest Mean of Exemplars rule. \medskip

\noindent Venkatesan \etal \cite{gan1} demonstrated that Generative Adversarial Networks can be used to as a privacy preserving alternative to iCarl. They propose to store a separate generator at each increment, which can consume a lot of storage. Our approach stores only one generator to generate representatives of all the previous increments. Furthermore, they do not deal with the bias introduced during distillation.\medskip

\noindent Wu \etal \cite{neu} also empirically studied the use of Generative Adversarial Networks for incrementally learning the classifier. They proposed a way to remove the bias caused by imbalance in the number of samples in the dataset by introducing a scalar multiplier whose value is estimated using a validation set. Their approach, however, is still biased as we will discuss in Section \ref{modeldistbias}. \medskip

\noindent Another direction of research commonly found in literature trying to solve the problem of catastrophic forgetting involves freezing of certain weights in neural network. Xiao \etal \cite{expand1} propose to learn a model structured like a tree that expands as it encounters more classes. Rusu \etal \cite{pnn} propose progressive neural networks, wherein they expand each layer horizontally. Kirkpatrick \etal \cite{ewc} propose Elastic Weight Consolidation (EWC), a regularization scheme to slow down learning on weights that are important to previous tasks. EWC builds a Fisher matrix using training data to determine which weights it should regularize.\medskip

\noindent Kemker \etal \cite{fearnet} propose FearNet, a brain inspired model for incremental learning which does not require storing old examples. It consists of a dual memory system, where one is for short-term memories and the other for long-term memory. However, instead of learning the representation, they use pretrained ResNet embeddings to obtain features to be fed into their model.\medskip

\noindent Lopez-Paz \& Ranzato propose Gradient Episodic Memory \cite{gem} where they solve forgetting problem by storing data from previous training sessions and performing gradient update such that the error on previously learned tasks do not increase. PathNet, proposed by Fernando \etal \cite{pathnet} uses a genetic algorithm to find the optimal path through the network for each training session. For each training session, only optimal path through the network will be trainable, preventing any new training sessions from causing the model to lose knowledge about previous training data.


\section{Model Distillation} \label{modeldist}
In this section we formally describe and discuss the approach for incremental learning by distilling knowledge from an old classifier as is common in the literature \cite{gan1} \cite{neu}. This approach is applicable to any GAN, i.e. we are not limited to GANs which have auxiliary classifier in them. We assume that we have a source from which we receive real datasets i.e. $\mathcal S=\{\mathcal D_1, \mathcal D_2,\dots, \mathcal D_i\}$ Each dataset represents one increment of learning and contains tuples of training samples $x_i$ and their true labels $y_i$ i.e. $\mathcal D_i=\{(x_i, y_i)\}$. We only have a limited amount of observations in datasets obtained from $\mathcal S$. In contrast, we can generate a very large amount of training data from the generator at almost no cost. 

\smallskip
\begin{figure}[t]
	\begin{algorithm}[H]
		\KwIn{Dataset $\mathcal D$, Classifier $\phi$, Generator $\mathcal G$ and Temperature T}
		\KwOut{Classifier $\phi$ and Generator $\mathcal G$\\}
		\If{old Classifier $\phi$ exists}
		{Use generator $\mathcal G$ to generate $(x_i, y_i)$ and append them into $\mathcal D$}
		
		Train a new generator $\mathcal G$ and discriminator $\psi$ on $(x_i, y_i) \sim \mathcal D$\\
		\uIf{old Classifier $\phi$ exists}
		{$\mathcal D_{soft} = (x, y^{\phi(T)}) \ \forall \ x \in \mathcal D$ \\
			Train $\phi$ by minimizing $\mathbb{E}_{(x,y)\sim \mathcal D,\ (\,\cdot\, ,y')\sim 
				\mathcal D_{soft}}[\alpha * \mathcal H(y,\,\phi(x)) + (1-\alpha) * \mathcal H(y',\,\phi(x))]$}
		\Else
		{Train $\phi$ by minimizing $\mathbb{E}_{(x, y)\sim \mathcal D}[\mathcal H(y,\,\phi(x))]$}
		
		\caption{Incremental Knowledge Transfer Using Model Distillation}
	\end{algorithm}
\end{figure} 
\smallskip

\noindent The proposed setup consists of a classifier $\phi$ to which the knowledge has to be transferred,
a generator $\mathcal G$ able to generate training samples and a discriminator $\psi$.
Throughout the paper, we will use the notation $y_{i}^{M(t)}$ to denote the label of $x_i$ generated 
using model $M$ with temperature $t$, $y_i$ to denote the true label of $x_i$ and $\mathcal H$ to denote 
the Cross Entropy. Whenever we receive a new real dataset from $\mathcal S$, it is fed along 
with other prerequisites to Algorithm 1, which consists of three phases as discussed below:

\smallskip\noindent {\bf Generate and add samples to dataset using the latest generator} { (Steps 1 - 3) }\\
The main goal of this step is to generate samples belonging to each class present in the previous datasets.
Therefore, this step is skipped if there are no previous datasets i.e. the classifier has not 
been trained yet. We generate a fixed amount $k$ samples for each class. In  case the generator
$\mathcal G$ is not conditional, we generate $k*m$ samples where $m$ is the number of classes
the classifier $\phi$ has been trained on. These samples are then appended into dataset $\mathcal D$. 
We discard $\mathcal G$ after this step since there is no further use of it.

\smallskip \noindent {\bf Train generator $\mathcal G$ and discriminator $\psi$ on samples
	obtained from $\mathcal D$} (Step 4)\\
Several different types of GANs can be used in this phase. The goal at this phase is to train $\mathcal G$ 
to be used as a privacy preserving alternative to storing $\mathcal D$ at the cost of reduced 
sample quality. This approach keeps at most one $\mathcal G$ and $\psi$ stored at a time, and they are both 
reinitialized at every increment instead of being trained continuously because several variants of GANs 
suffer from mode collapse when trained for a longer periods of time.

\smallskip \noindent {\bf Transfer previous knowledge to classifier $\phi$ using Model-Distillation} (Steps 5 - 9)\\
If there is no old classifier to transfer the knowledge from, then we simply minimise the cross-entropy between the real and the predicted targets. Otherwise, we generate a dataset $D_{soft}$ with softer targets $y^{\phi(T)}$ using temperature $T$. The classifier $\phi$ from which we are obtaining the soft targets at this point is trained only on classes belonging to older datasets. In the next step, we proceed by training the classifier $\phi$ on new classes by using standard weighted distillation loss. In first part of the loss, we calculate the cross entropy between the predicted target $\phi(x)$ and real target $y$ obtained from $\mathcal D$. In second part, we calculate cross entropy between the predicted target $\phi(x)$ and soft target $y^{\phi(T)}$ obtained from $\mathcal D_{soft}$. $\alpha$ is the weight constant used to determine the weight of these two terms. It is necessary to train on real targets in addition to softened targets as the classifier used to generate the soft targets had no knowledge about the new classes, so omitting it would result in inability to learn about them.

\section{Model Distillation Trains a Biased Classifier} \label{modeldistbias}
When we perform distillation using trained classifier $\phi$, the resulting model can be biased in three different ways:
\begin{enumerate}
	\itemsep 0.5em
	\item The number of samples generated by $\mathcal G$ can be different than the number of samples present per unique class in $\mathcal D$ i.e $k*m >> |\mathcal D|$ or $k*m << |\mathcal D|$ where $k$ is the number of generated samples per class and $m$ is the number of unique classes in $\mathcal D$. 
	\item When distillation is performed on new classes the soft targets obtained will indicate that new classes have high similarity to older classes, even if they are not similar. This introduces a bias against the new classes.
	\item When distillation is performed on old classes, the soft targets will indicate that the old classes have no similarity to new classes, even if they are similar. This introduces a bias against the old classes.
\end{enumerate}
We illustrate these biases through a simple example. We assume that we have an ideal softmax based image classifier $\mathcal \phi_{ideal}$ that can classify the samples it was trained on with zero error. For the second type of bias, we assume that $\mathcal \phi_{ideal}$ is already trained on classes $\mathcal D=(Buildings, Boats)$. If this $\mathcal \phi_{ideal}$ has to be trained on $\mathcal D=(Cats)$, the second term of equation in step 7 will of Algorithm 1 will indicate that the Cat class images have a high similarity to the images of Buildings and Boats, which is clearly not the case. This introduces a bias against the Cat class as their class probabilities are incorrect. This happens because the soft labels $y'$ are obtained from a classifier that was trained only on the old classes i.e. $D=(Buildings, Boats)$.

For the third type of bias, we assume that $\mathcal \phi_{ideal}$ is already trained on classes $\mathcal D=(Dogs, Ocean)$. If this $\mathcal \phi_{ideal}$ has to be trained on $\mathcal D=(Cats)$, the second term of equation in step 7 will of Algorithm 1 will indicate that the Dog images have zero similarity to the images of Cats. This introduces a bias against the Dog class as its class probabilities are incorrect. Which is also caused because the soft labels $y'$ are obtained from a classifier that was trained only on the old classes.

The effect of these biases can be slightly mitigated by carefully selecting the parameters $k$ and $\alpha$, however, we present an alternative solution to this problem by using an auxiliary classifier to mitigate this bias.

\smallskip
\begin{figure}[t]
	\begin{algorithm}[H]
		\SetAlgoLined
		\KwIn{Dataset $\mathcal D$, Classifier $\phi$, Generator $\mathcal G$ and Temperature T}
		\KwOut{Classifier $\phi$ and Generator $\mathcal G$}
		\If{old Classifier $\phi$ exists}
		{Use generator $\mathcal G$ to generate $(x_i, y_i)$ and append them into $\mathcal D$}
		
		Train a new generator $\mathcal G$ and discriminator $\psi$ on $(x_i, y_i) \sim \mathcal D$\\
		\uIf{old Classifier $\phi$ exists}
		{$\mathcal D_{soft} = (x, y^{\mathcal\psi_{{\mathcal{AC}}}(T)})\ \forall \ x \in \mathcal D$\\
			Train $\phi$ by minimizing $\mathbb{E}_{(x, y)\sim \mathcal D_{soft}}[ \mathcal H(y,\,\phi(x))]$}
		\Else
		{Train $\phi$ by minimizing $\mathbb{E}_{(x, y)\sim \mathcal D}[ \mathcal H(y,\,\phi(x))]$}
		
		\caption{Incremental Knowledge Transfer Using AC-Distillation}
	\end{algorithm}
\end{figure}
\smallskip

\section{AC-Distillation} \label{acdist}
The AC-Distillation (Auxiliary Classifier Distillation) is outlined in algorithm 2. The key idea of AC-Distillation is to generate $\mathcal D_{soft}$ by distilling knowledge from auxiliary classifier $\psi_{\mathcal{AC}}$ of GAN instead of the old classifier $\phi$. The details of differences of this approach are described below:
\begin{enumerate}
	\item In step 6, the distillation is done using the auxiliary classifier $\psi_{\mathcal{AC}}$ obtained from GAN that we just trained. Therefore, $\psi_{\mathcal{AC}}$ contains information about the new classes present in $\mathcal D$ in addition to old classes.
	\item In step 7, we train the classifier $\phi$ on $\mathcal D_{soft}$ that we generated in step 6. Note that we calculate cross-entropy only using the soft targets present in $\mathcal D_{soft}$ and do not use the real targets present in $\mathcal D$. This is because in our experiments, when using a source of distillation that contains information about all the classes that the classifier $\phi$ is to be trained on, using only $D_{soft}$ performed at least as good as using $\mathcal D_{soft} \cup \mathcal D$. Furthermore, this allows us to get rid of the parameter $\mathcal \alpha$, which reduces our hyperparameter search space.
\end{enumerate}
Note that in this method, we are limited to GANs that have an auxiliary classifier built into them. One example of of a GAN that satisfies this property is AC-GAN \cite{acgan}.\\
One might doubt about the usefulness of separately training a new classifier when there already exists a classifier in the GAN. However, this constraints the size and type of architectures that can be used as a classifier, as the auxiliary classifier is a part of the GAN, which has to be built in a certain way.


\section{Experiments} \label{experiments}
\subsection{Benchmark}
We use the benchmark protocol suggested by \cite{icarl}, that is, we train using each method
in a class incremental way, where the classifier $\mathcal \phi$ receives a new set of previously unseen classes each increment. After each increment, the classifier $\mathcal \phi$ is evaluated on the test set consisting of new and old classes combined. We use MNIST \cite{mnist} and CIFAR10 \cite{cifar10} as real sources $\mathcal S$. Each source $\mathcal S$
contains 5 datasets $\mathcal D$ where each dataset $\mathcal D$ consists of 2 randomly picked previously unseen classes.

\begin{figure}[t]
	\centering
	\begin{tabular}{cc}
		\includegraphics[width=6cm]{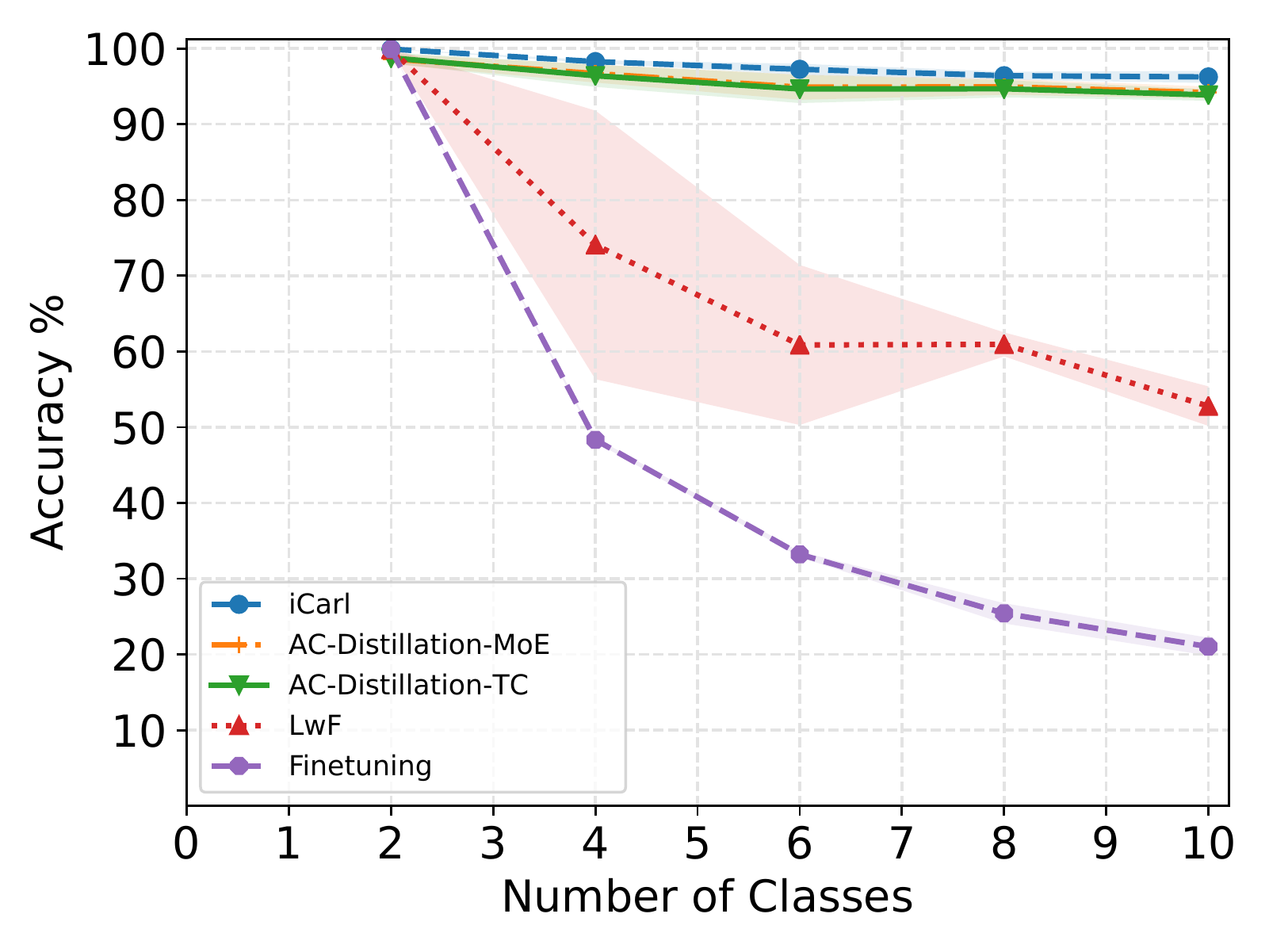}&
		\includegraphics[width=6cm]{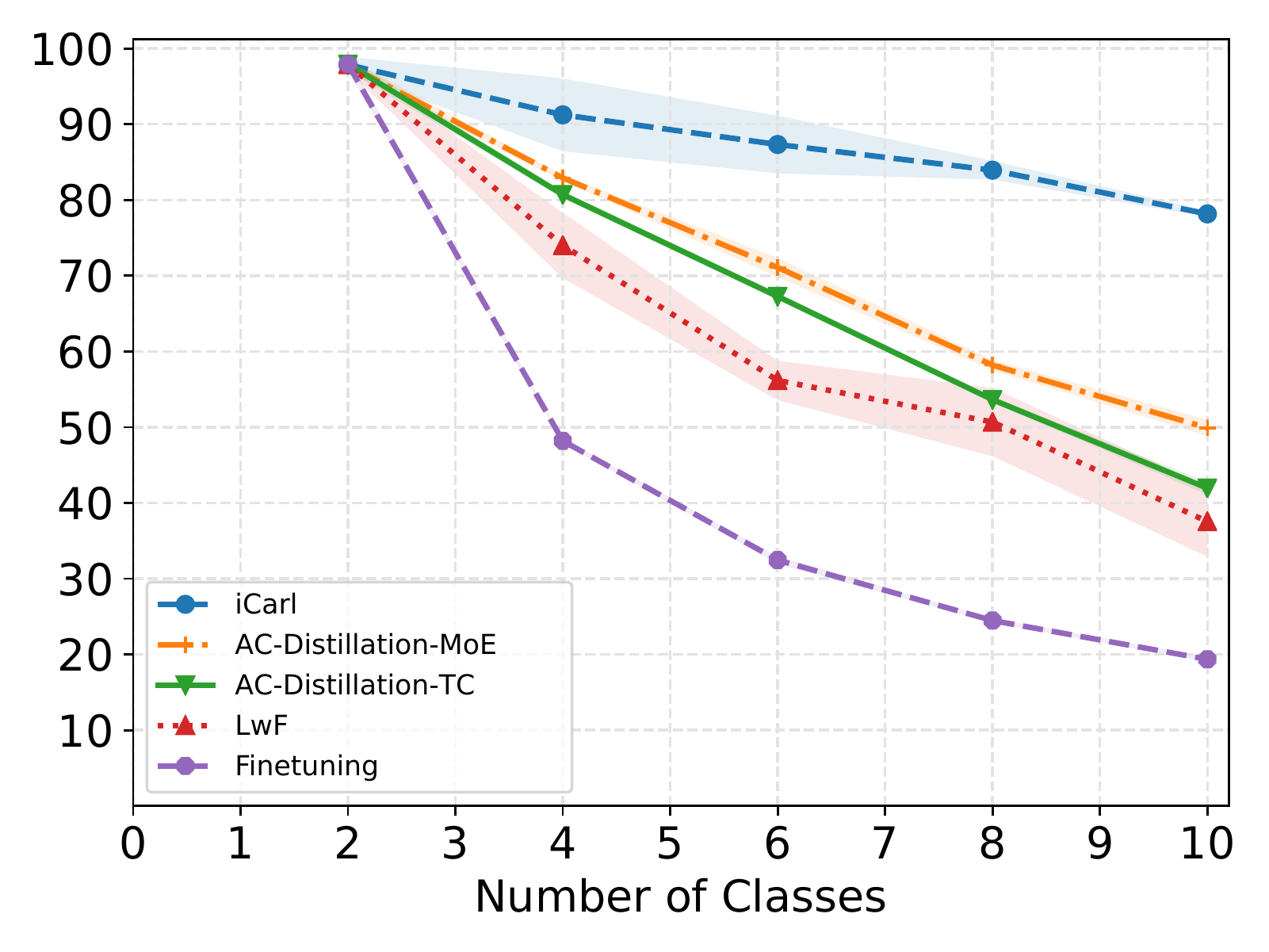}
		\\ (a) MNIST &(b) CIFAR10
	\end{tabular}
	\caption{All experiments were repeated 3 times. Mean and standard
		deviation are reported. (a) Results on MNIST with 2 classes per dataset $\mathcal D$. Here the performance AC distillation approaches is comparable to iCarl. (b) Results on CIFAR10 with 2 classes per dataset $\mathcal D$. Here the performance of AC distillation approaches is worse than iCarl. This is because the AC-GAN is unable to properly learn the dataset. In contrast, iCarl is storing the real images as exemplars. In both datasets, AC distillation has the best performance among the privacy preserving approaches.} \label{fig:results}
\end{figure}

\subsection{Implementation details} \label{impl}
\noindent{\bf Classifier:} We train Resnet-32 \cite{resnet} with batch size of 100. Learning rate is set as 2 and decayed by 0.2. Sigmoid is used in the final layer and binary cross entropy is used as loss function. For MNIST, we train for 15 epochs and decay the learning rate after 8 and 12 epochs. For CIFAR10, we train for 70 epochs and decay the learning rate after 45, 60 and 68 epochs. 

\medskip\noindent{\bf AC-GAN:} We use the CIFAR10 architecture provided
by Odena \etal \cite{acgan} with a few minor changes in their hyperparameters. More specifically, we train MNIST for 20 epochs
each increment, decaying the learning rate at epochs 11 and 16 and CIFAR10 for 1000
epochs each increment, decaying the learning rate after 900 epochs. Learning rate is decayed by
0.1.

\subsection{Approaches compared} \label{approaches}
We compare the following techniques in our experiments:
\begin{enumerate}
	\itemsep0.5em 
	\item{iCarl \cite{icarl}:} Distillation using old classifier $\phi$ on herded and new examplars. Classification is done using the Mean of Exemplars rule. This approach is not privacy-preserving as old class data is stored.
	\item{LwF \cite{lwf}:} Distillation using old classifier $\phi$ only on new examples. Classification is done using newly trained classifier. This is a privacy-preserving approach.
	\item{Finetuning:} No distillation and performing classification using the newly trained classifier. Here we simply finetune the old classifier on newly arriving classes without taking any measures to prevent catastrophic forgetting.
	\item{AC-Distillation-TC:} Distillation using auxiliary classifier $\mathcal \psi_{\mathcal{AC}}$ 
	of GAN on generated and new examples. Classification is done using the newly trained classifier. This is a privacy-preserving approach.
	\item{AC-Distillation-MoE:} Distillation using auxiliary classifier $\mathcal \psi_{\mathcal{AC}}$
	of GAN on generated and new examples. Classification is done using Mean of Exemplars classification rule. This is a privacy-preserving approach.
\end{enumerate}

\subsection{Results}
\subsubsection{MNIST:}
We can see in Fig. \ref{fig:results}\color{red}a \color{black}that the performance of AC-Distillation-TC is resembles AC-Distillation-MoE. This shows that on simple datasets, the trained classifier can be used directly to obtain a good classification accuracy. When comparing to iCarl, the performance is slightly lower, which is due to the limitation of AC-GAN to perfectly model the dataset. However, if only the privacy-preserving approaches are compared, AC-Distillation approaches beat other approaches by a large margin. The finetuning approach forgets the existence of earlier classes.\\

\subsubsection{CIFAR10:} CIFAR10 is a much more complicated dataset to model as compared to MNIST. This is partly because the size of the image is much smaller when compared to contents of the image. We can see from Fig. \ref{fig:results}\color{red}b \color{black} that the performance of AC-Distillation methods is significantly worse than iCarl, which is due to the inability of the AC-GAN to properly learn the dataset. However, even with the AC-GAN performing poorly on this dataset, AC-distillation methods perform better than the other privacy-preserving approaches.

If we compare performance of the two AC-distillation methods, we see that AC-Distillation-TC performs slightly worse than AC-Distillation-MoE. This shows that on more complicated datasets, Mean of Exemplars classification rule should be used to achieve the best performance as it slightly compensates for the inability of GANs properly model the dataset. \\

\noindent The results on MNIST indicate that AC-Distillation is a viable approach to privacy-preserving incremental learning. However, when using more complicated datasets, AC-GAN is unable to properly model the dataset. This results in a poor performance when using AC-distillation on such datasets. In order to achieve a performance comparable to non privacy-preserving approaches on more difficult datasets, AC-GAN has to be replaced with more advanced GAN architectures. 

\begin{figure}[t]
	\centering
	\begin{tabular}{c}
		\includegraphics[width=8cm]{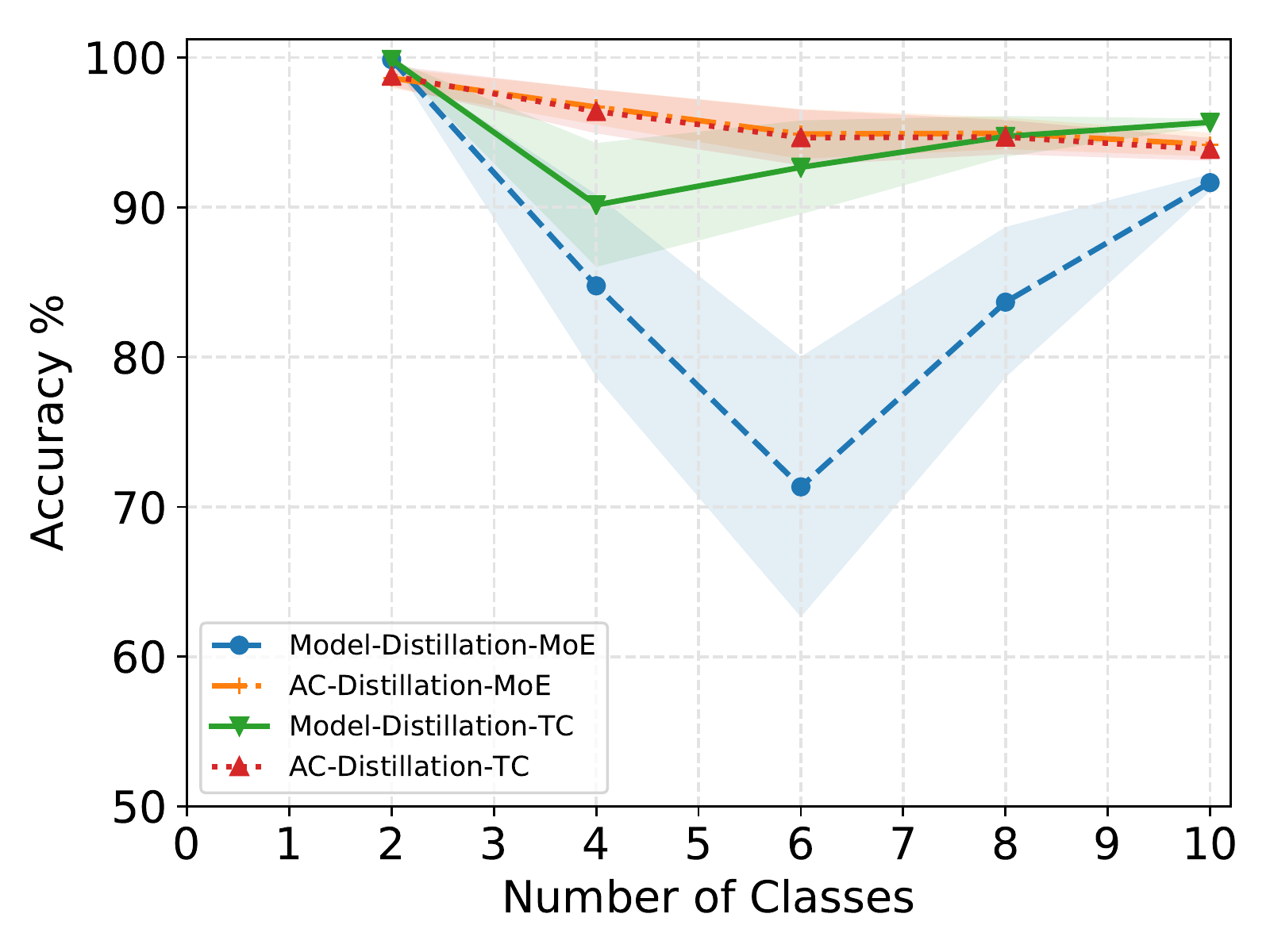}
	\end{tabular}
	\caption{Comparison of AC-Distillation methods with Model distillation on MNIST using 2 classes per
		dataset $\mathcal D$. All experiments were repeated 3 times. Mean and standard deviation are reported. Y-axis is truncated to highlight the differences. We can see that the AC distillation approaches have lower variance and perform significantly better than the Model distillation approaches.}
	\label{fig:acvsm}
\end{figure}

\setlength{\tabcolsep}{0pt}
\begin{figure}
	\centering
	\begin{tabular}{c@{\hspace{-0.5cm}}c@{\hspace{-0.5cm}}c@{\hspace{-0.5cm}}c@{\hspace{-0.5cm}}cc}
		\includegraphics[width=2.7cm]{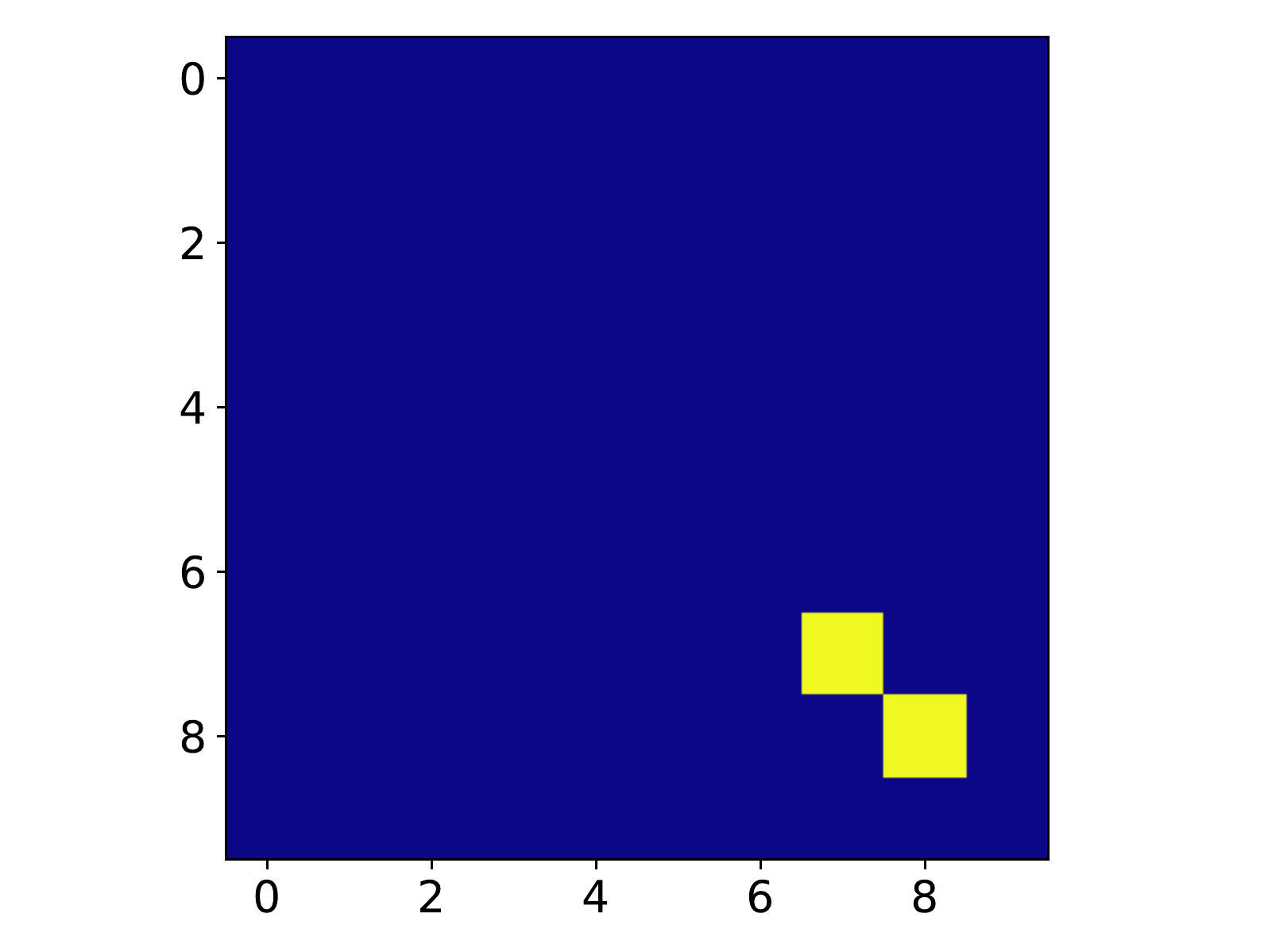}&
		\includegraphics[width=2.7cm]{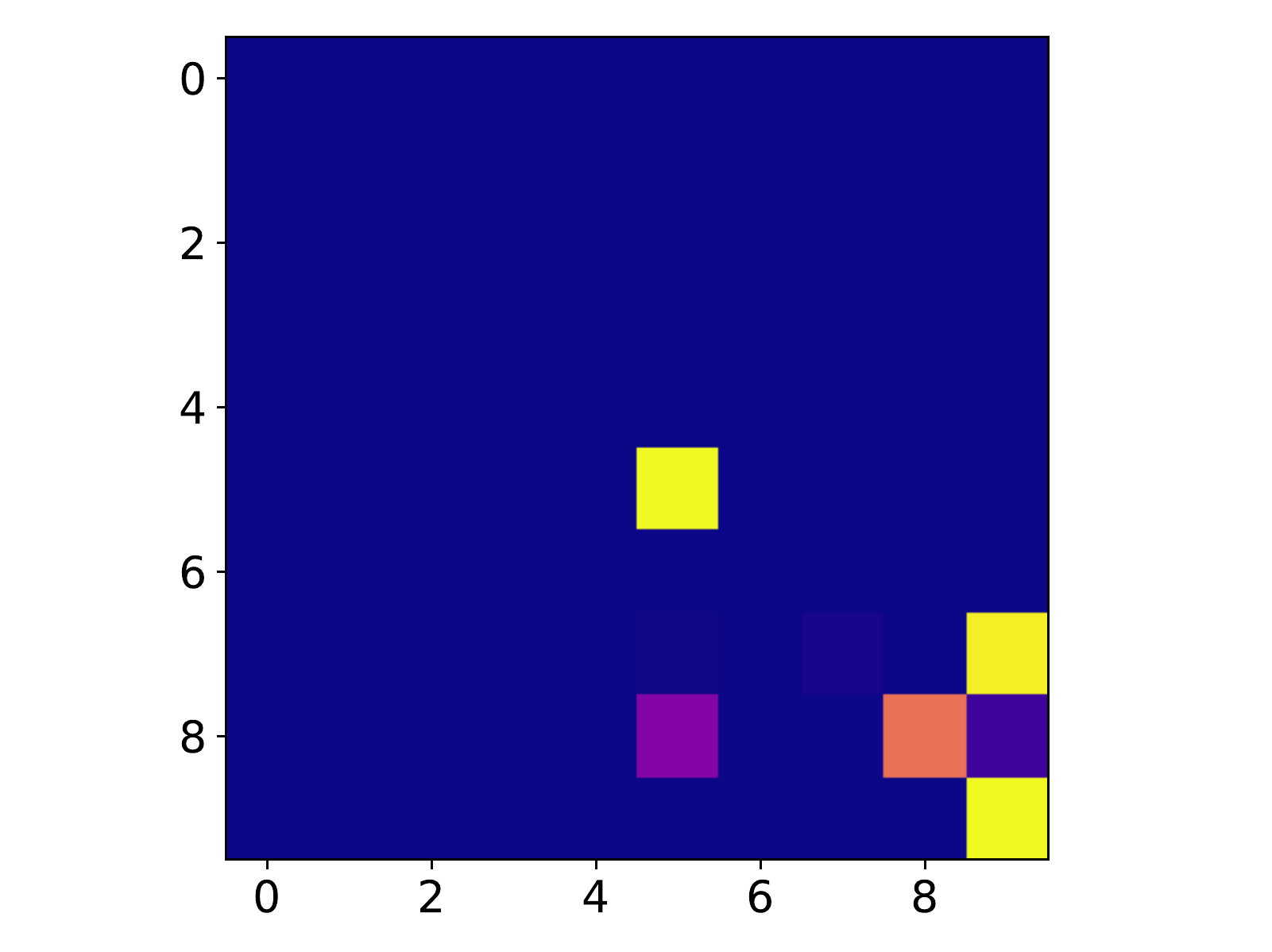}&
		\includegraphics[width=2.7cm]{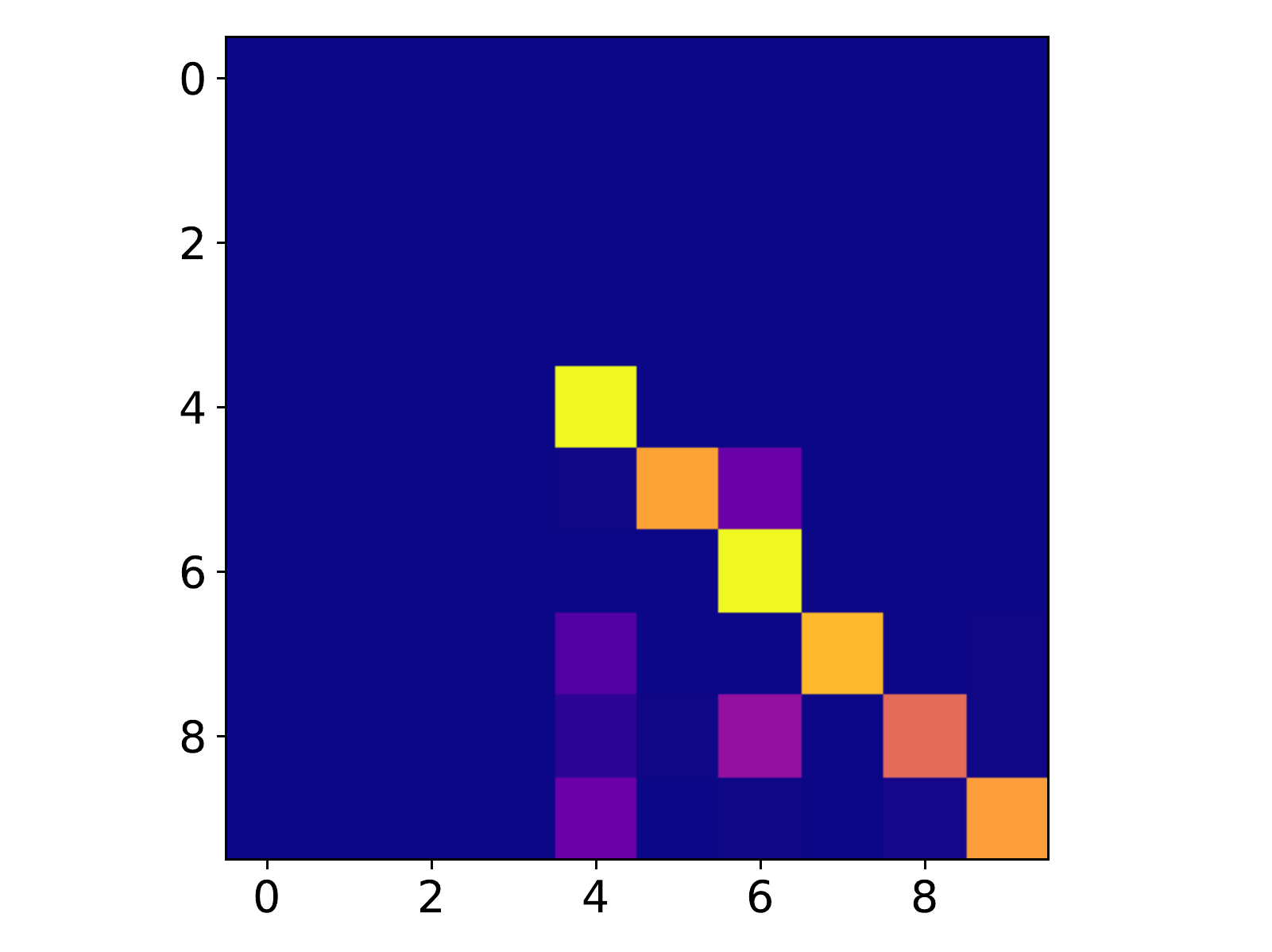}&
		\includegraphics[width=2.7cm]{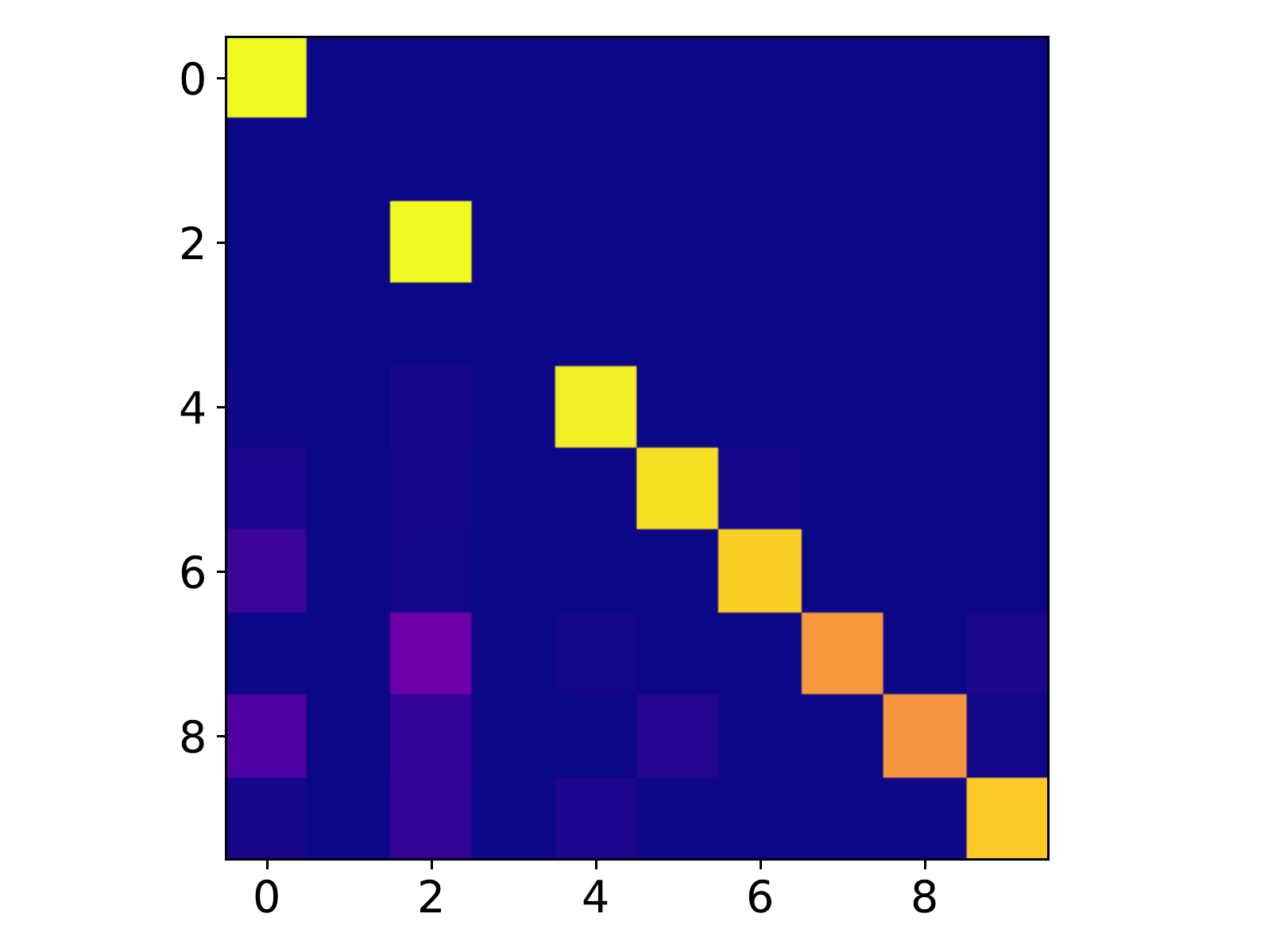}&
		\includegraphics[width=2.7cm]{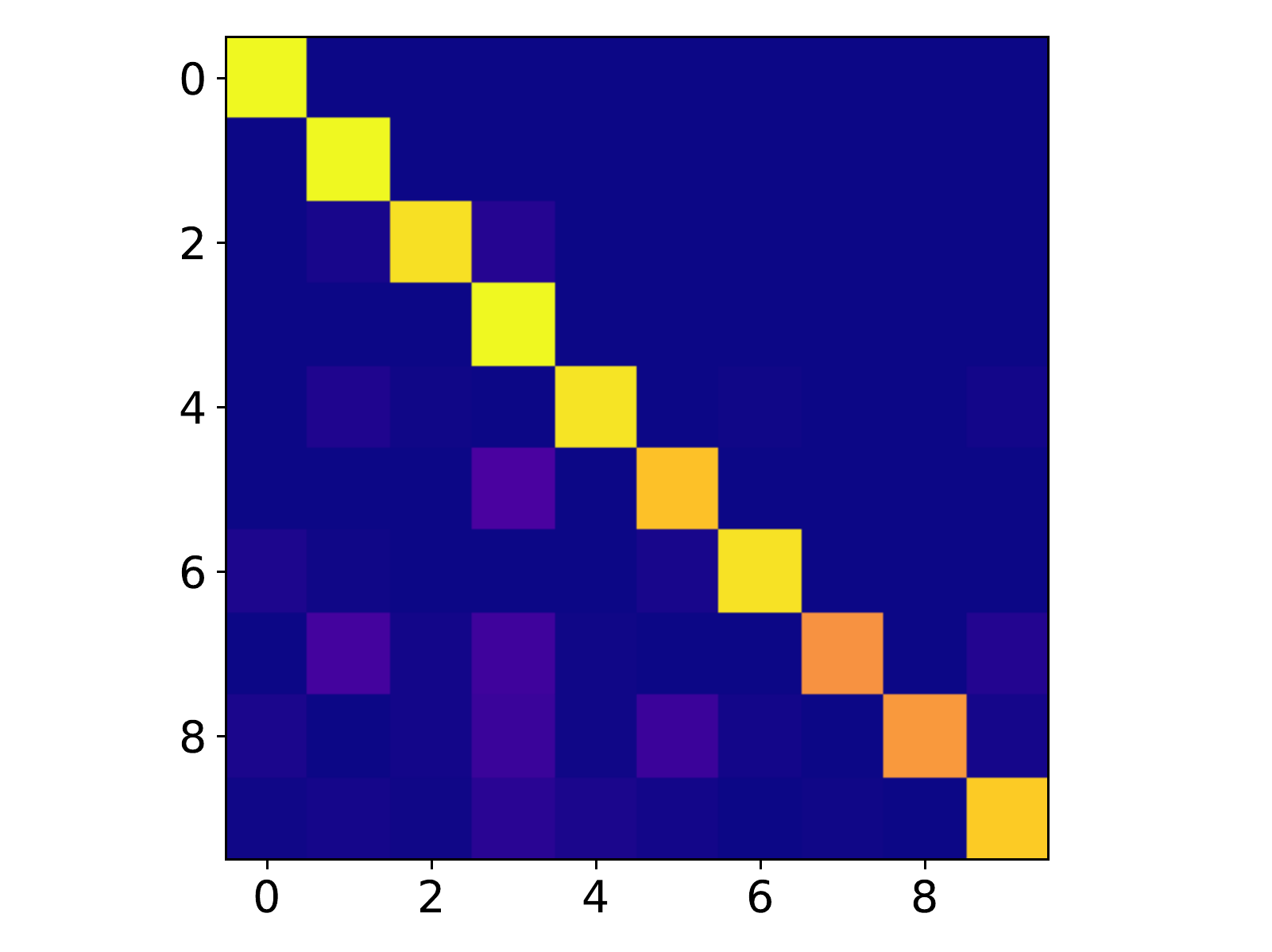}&
		\multirow{-5.3}{*}{
			\includegraphics[height=3.75cm]{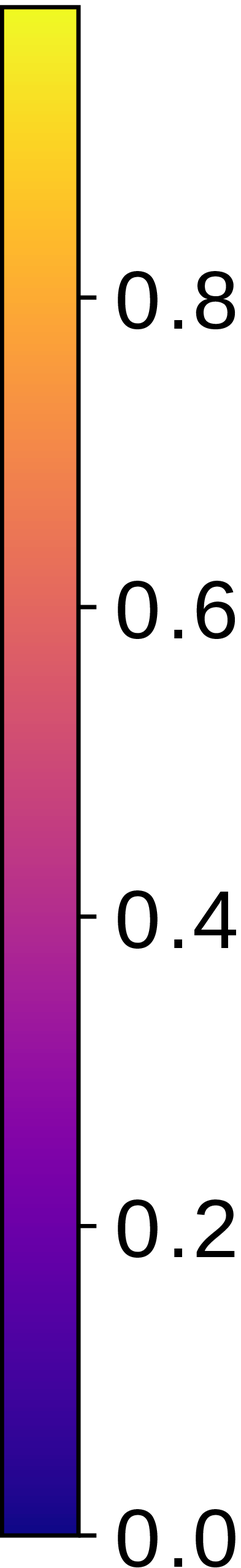}}\\[-1.5ex]
		\includegraphics[width=2.7cm]{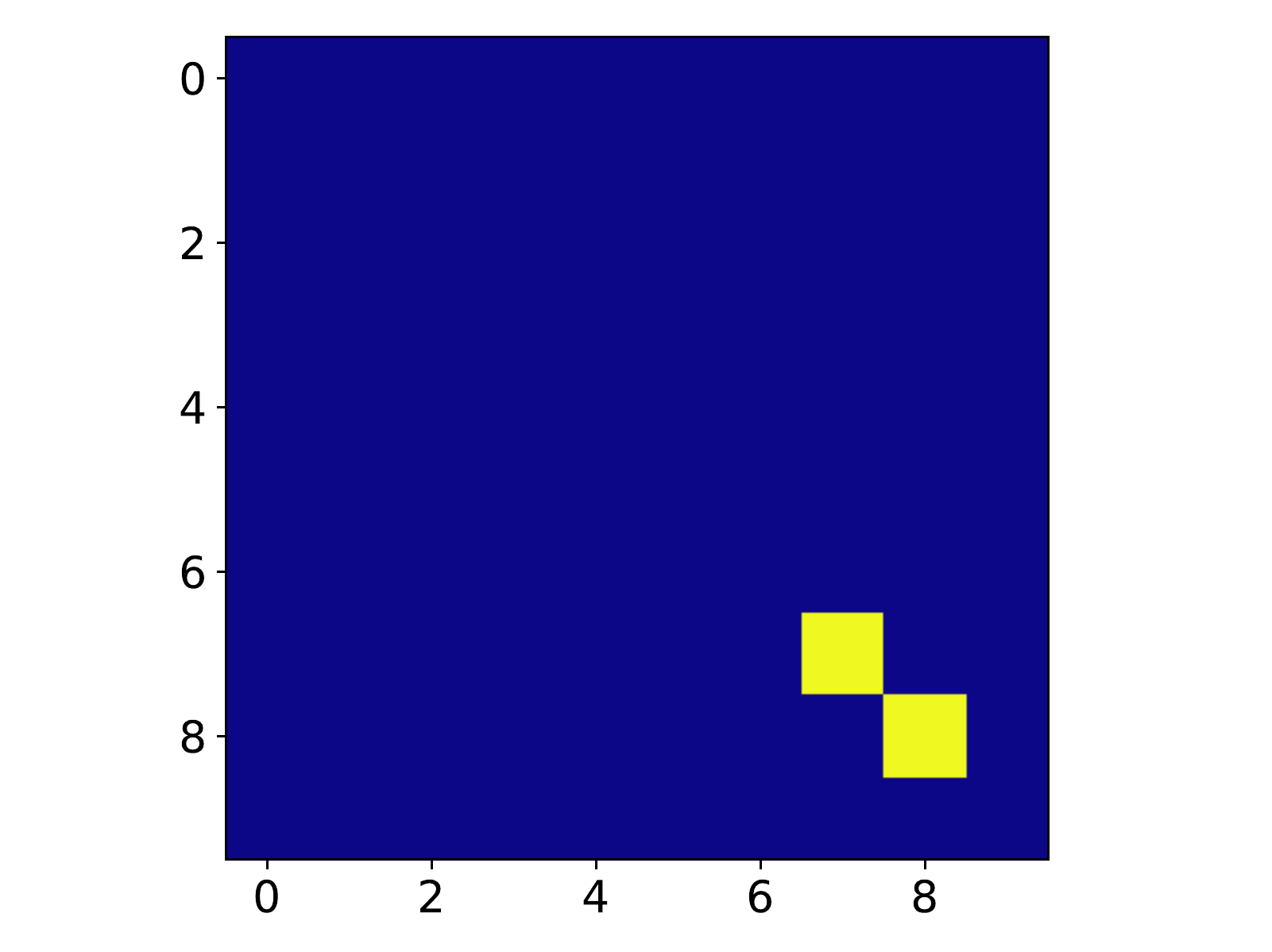}&
		\includegraphics[width=2.7cm]{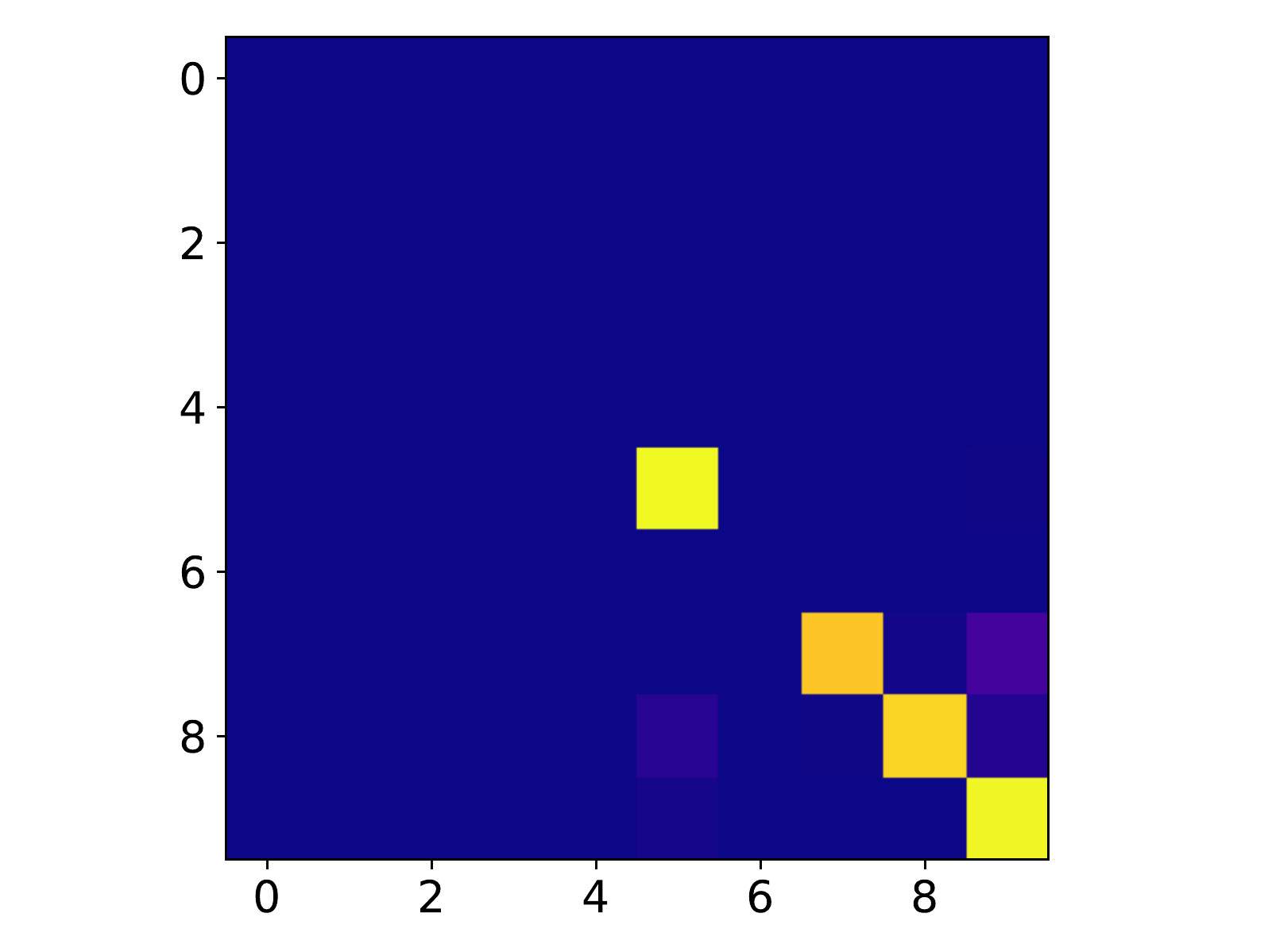}&
		\includegraphics[width=2.7cm]{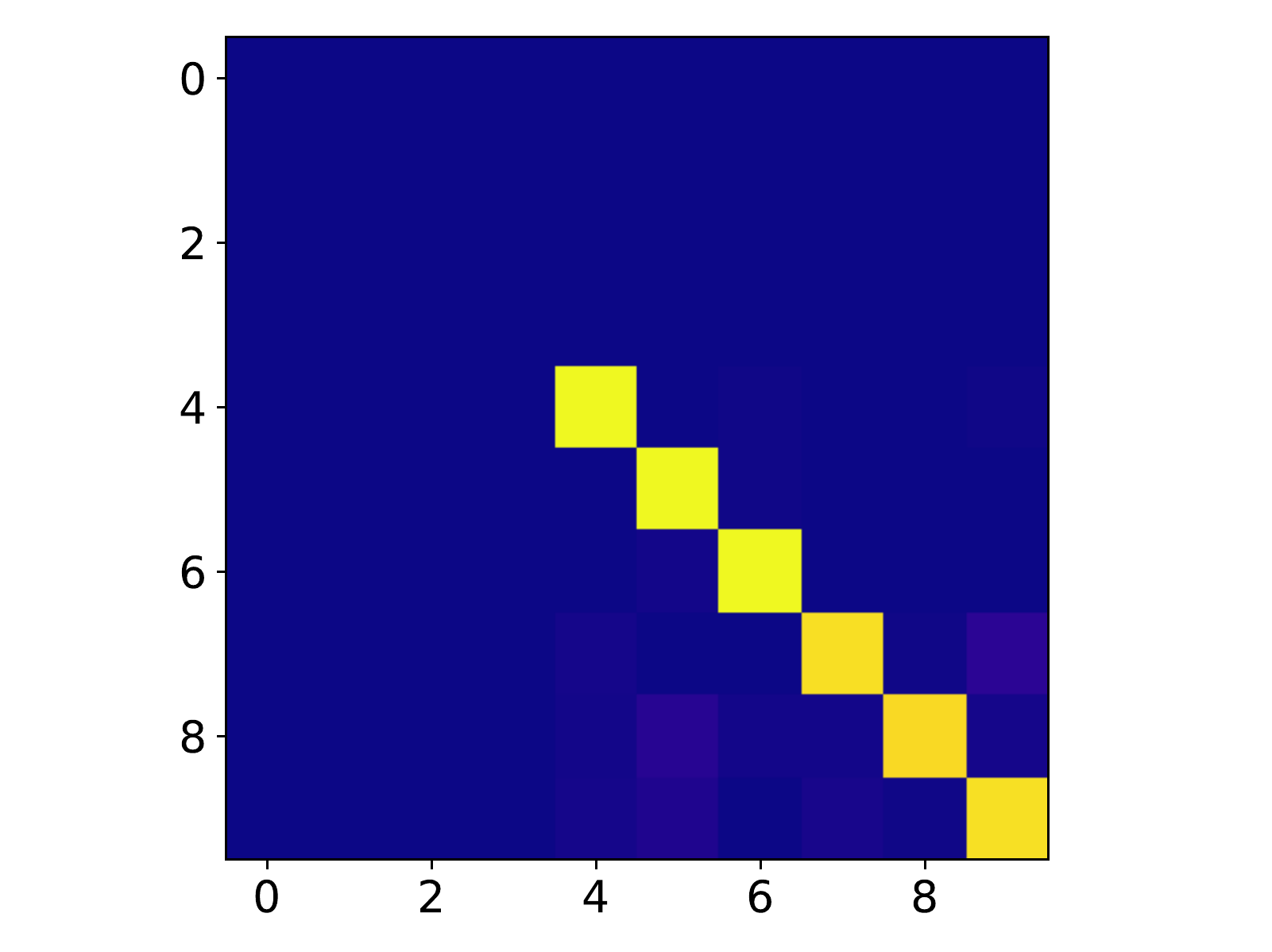}&
		\includegraphics[width=2.7cm]{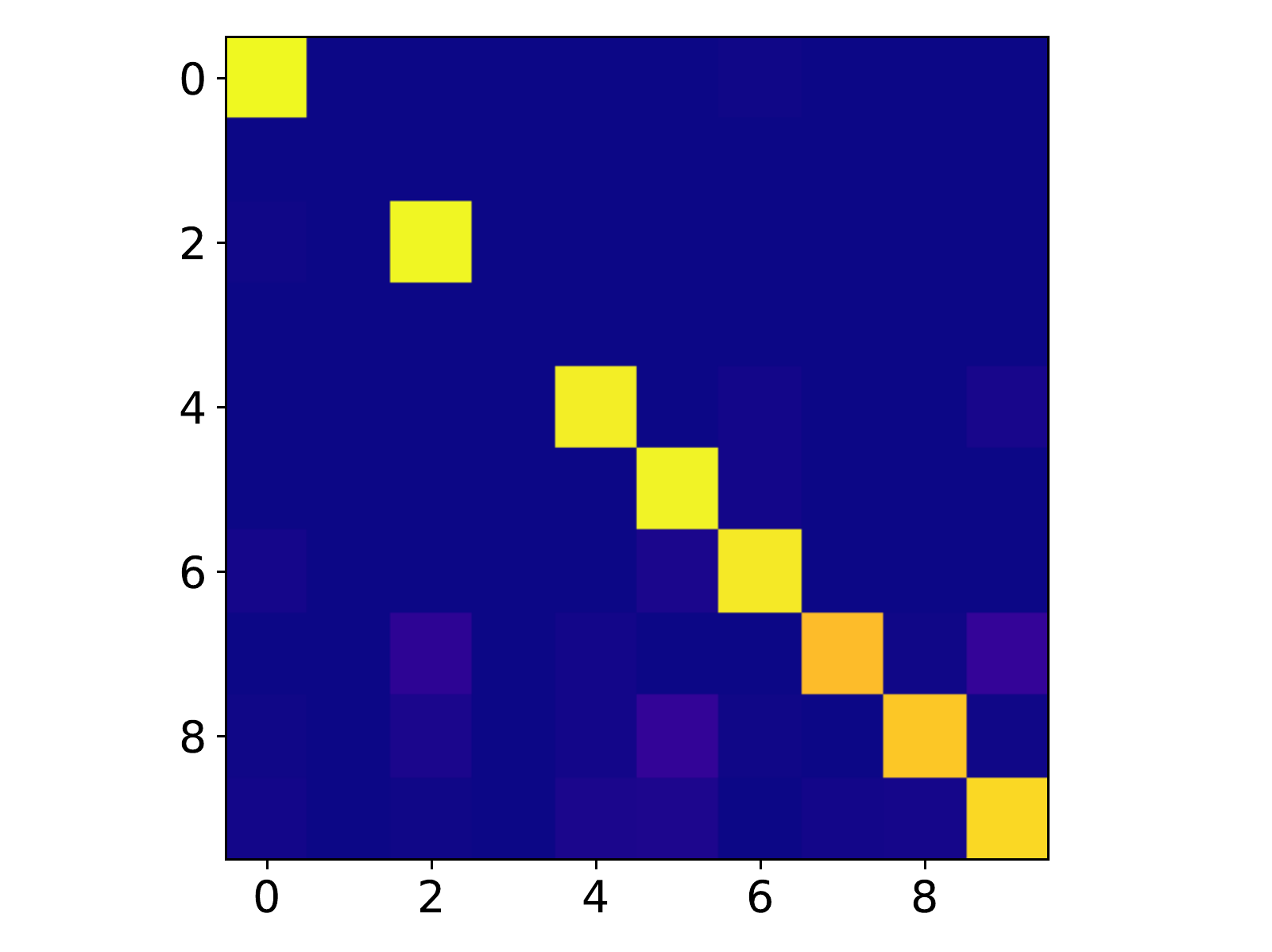}&
		\includegraphics[width=2.7cm]{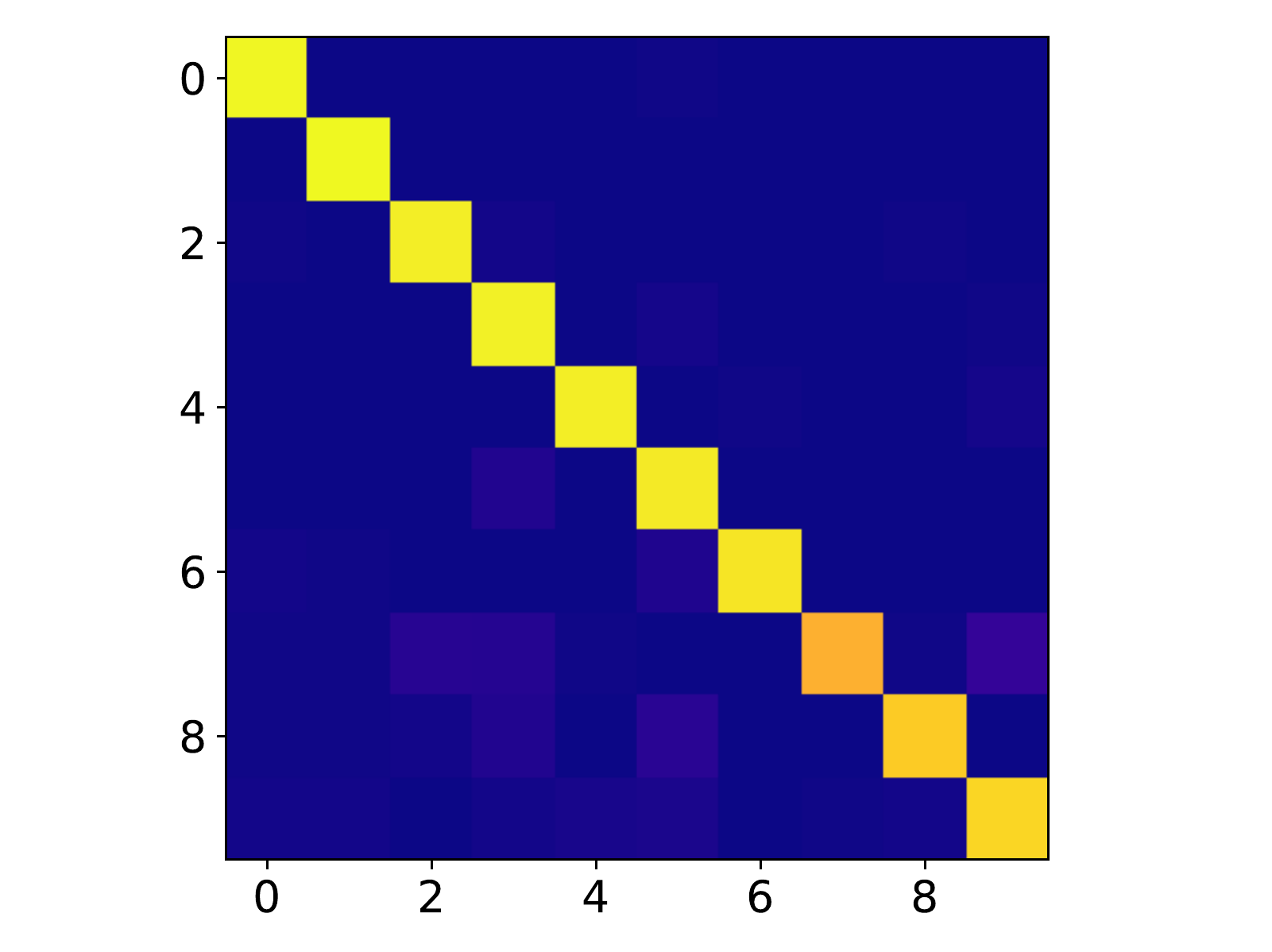}&\\
	\end{tabular}
	\caption{Confusion matrices for incremental training session on MNIST where x-axis denotes the ground truth and y-axis denotes the
		predicted label. Top: Confusion matrices for Model-Distillation-MoE. Bottom: Confusion matrices
		for AC-Distillation-MoE. These matrices are for a single run of each experiment. Each matrix indicates a separate training session with new knowledge added to it as we go towards the right side. The new classes in the above example arrived in the this order: $\mathcal S=\{(7,8),\,(5,9),\,(4,6),\,(0,2),\,(1,3)\}$. We can observe from these figures that the Model distillation approach tends to forget the knowledge about the older classes in the middle increments while the AC-distillation approach is able to retain the knowledge about old classes.}
	\label{fig:acvsm2}
\end{figure}
\subsection{Comparison of AC-Distillation with Model Distillation}
In this section, we compare Auxiliary Classifier based knowledge distillation methods with Model distillation methods. More specifically, we compare {\it AC-Distillation-TC} and {\it AC-Distillation-MoE} described in Section \ref{approaches} with the following:
\begin{enumerate}
	\itemsep0.5em 
	\item{ Model-Distillation-TC:} Distillation using old classifier $\phi$ on generated and new examples. Classification is done using the newly trained classifier.
	\item{ Model-Distillation-MoE:} Distillation using old classifier $\phi$ on generated and new examples. Classification is done using the Mean of Exemplars classification rule.
\end{enumerate}

To make the comparison fair, all the experiments were run using AC-GAN and the parameters described in Section \ref{impl}. The results of the comparison are shown in Fig. \ref{fig:acvsm}. At each increment of training, the classifier receives two new classes to train on. It can be seen that the Model distillation methods have higher variance and perform worse as compared to Auxiliary Classifier based distillation methods. The decreased performance of Model distillation methods is caused due to the biases discussed in Section \ref{modeldistbias}.

To understand their behavior, we can see the comparison of confusion matrices for Auxiliary Classifier and Model distillation based Mean of Exemplar classification methods in Fig.~\ref{fig:acvsm2}. It appears that when using AC-Distillation, the 
Mean of Exemplars classification rule is able to classify both the new and old classes properly. However,
in the case of Model distillation the Mean of Exemplars rule is unable to retain the knowledge about the old classes properly, and incorrectly classifies them as members of the new classes. As more and more new classes are fed to the classifier to learn about, the performance of Model distillation methods becomes comparable with AC-distillation methods. \\

From the above analysis, we can observe that when we have a small number of classes to incrementally train on, the effect of bias is significant in Model distillation methods. This leads to poor performance of Model distillation methods when compared to AC-distillation methods. However, when the number of classes is increased, the effect of bias is mitigated to some extent and performance of the classifier becomes comparable with AC-distillation methods. In contrast to this, AC-distillation methods do not suffer from this bias at any number of classes.


\section{Conclusion} \label{conclusion}
We have shown that performing knowledge distillation from the auxiliary classifier of AC-GAN corrects the biases that are prevalent in GAN based incremental learning methods. We proposed the AC-distillation approach to correct the biases. On MNIST, we studied the biases in model distillation approaches and demonstrated the effectiveness of the AC-distillation approach. On more complicated datasets, the performance is significantly worse due to the inability of AC-GAN to properly model the dataset. However, if a more advanced GAN is used that can properly model the dataset, the performance of the proposed approach is comparable to state of the art rehearsal based incremental learning methods, while being privacy preserving.

\bibliographystyle{splncs03}

\end{document}